\PassOptionsToPackage{table}{xcolor}

\documentclass[]{FudanCVL}

\usepackage{silence}
\usepackage[misc]{ifsym}

\usepackage{amsmath,amsfonts,amssymb}
\usepackage{pifont}
\usepackage{enumitem}
\usepackage{tabularx}
\usepackage{threeparttable}
\usepackage{xspace}

\definecolor{mygray}{gray}{.90}
\definecolor{myrowgray}{RGB}{240,240,240}
\definecolor{myRed}{HTML}{F14F4A}
\definecolor{myBlue}{HTML}{000099}
\newcommand{\pub}[1]{\color{black}{\scriptsize{[{#1}]}}}
\newcommand{\xmark}{\textcolor{lightgray}{\ding{55}}}
\newcommand{\cmark}{{\ding{51}}}
\newcommand{\etal}{\textit{et al.}\xspace}
\newcommand{\keywords}[1]{\par\vskip 0.15cm\noindent{\sffamily\bfseries Keywords:\ }#1}

\crefname{figure}{Fig.}{Figs.}
\crefname{table}{Tab.}{Tabs.}

\title{\vspace{-1pt}{\fontsize{19pt}{22.5pt}\selectfont Seek to Segment: Active Perception for Panoramic Referring Segmentation}}

\author{Song Tang}
\author{Shuming Hu}
\author{Xincheng Shuai}
\author{Henghui Ding}
\author{Yu-Gang Jiang}

\affiliation{Fudan University, China}

\abstract{
Existing referring segmentation models passively process static images captured from fixed perspectives, limiting their applicability in Embodied AI, where agents must perform active perception in the continuous 360$^\circ$ environments.
To bridge this gap, we introduce a novel task: \textbf{Active Panoramic Referring Segmentation~(APRS)}. In this setting, an agent is required to adjust its viewing direction~($\Delta\theta, \Delta\phi$) to explore the 360$^\circ$ environment, seeking the object specified by a user instruction for segmentation.
To tackle this challenging task, we propose \textbf{PanoSeeker}, a memory-augmented agent for efficient APRS.
Rather than relying on heuristic scanning, PanoSeeker integrates a Vision-Language Model~(VLM) with \textbf{EgoSphere}---an explicit spatial visual memory.
By progressively integrating sequential local observations into a unified 360$^\circ$ representation, EgoSphere enables the agent to plan efficient and non-redundant search trajectories.
Once the target is found, the agent performs active viewpoint alignment and outputs the segmentation mask.
Furthermore, we curate an expert-annotated search trajectory dataset with memory timelines for Supervised Fine-Tuning, followed by Reinforcement Learning post-training to explicitly optimize PanoSeeker's exploration efficiency.
Extensive experiments on our newly established APRS benchmark demonstrate that PanoSeeker achieves superior search efficiency and segmentation accuracy, significantly outperforming adapted state-of-the-art baselines.
}

\metadata[Website]{\url{https://henghuiding.com/APRS/}}
\metadata[Email]{henghui.ding@gmail.com}

\begin{document}
\maketitle

\section{Introduction}
Referring Segmentation~\cite{res_survey,tang2026rose, vlt, gres,grex,vltpami,MeViS} has achieved remarkable progress by segmenting target objects guided by natural language expressions. Powered by recent advancements in Vision-Language Models (VLMs)~\cite{llava,qwen2vl,qwen3vl,shuai2026psdesigner}, contemporary referring segmentation models~\cite{lisa,gsva,read,visionreasoner,pixelthink,MeViSv2,omniavs,RAVS} demonstrate exceptional reasoning ability for precise pixel-level segmentation. However, these methods passively process static images captured from fixed perspectives, which significantly limits their deployment in real-world Embodied AI~\cite{vln_survey,reverie}, where agents must actively perceive their surroundings to find targets within continuous 360$^\circ$ environments.

\begin{figure}[t]
    \centering
    \includegraphics[width=1\linewidth]{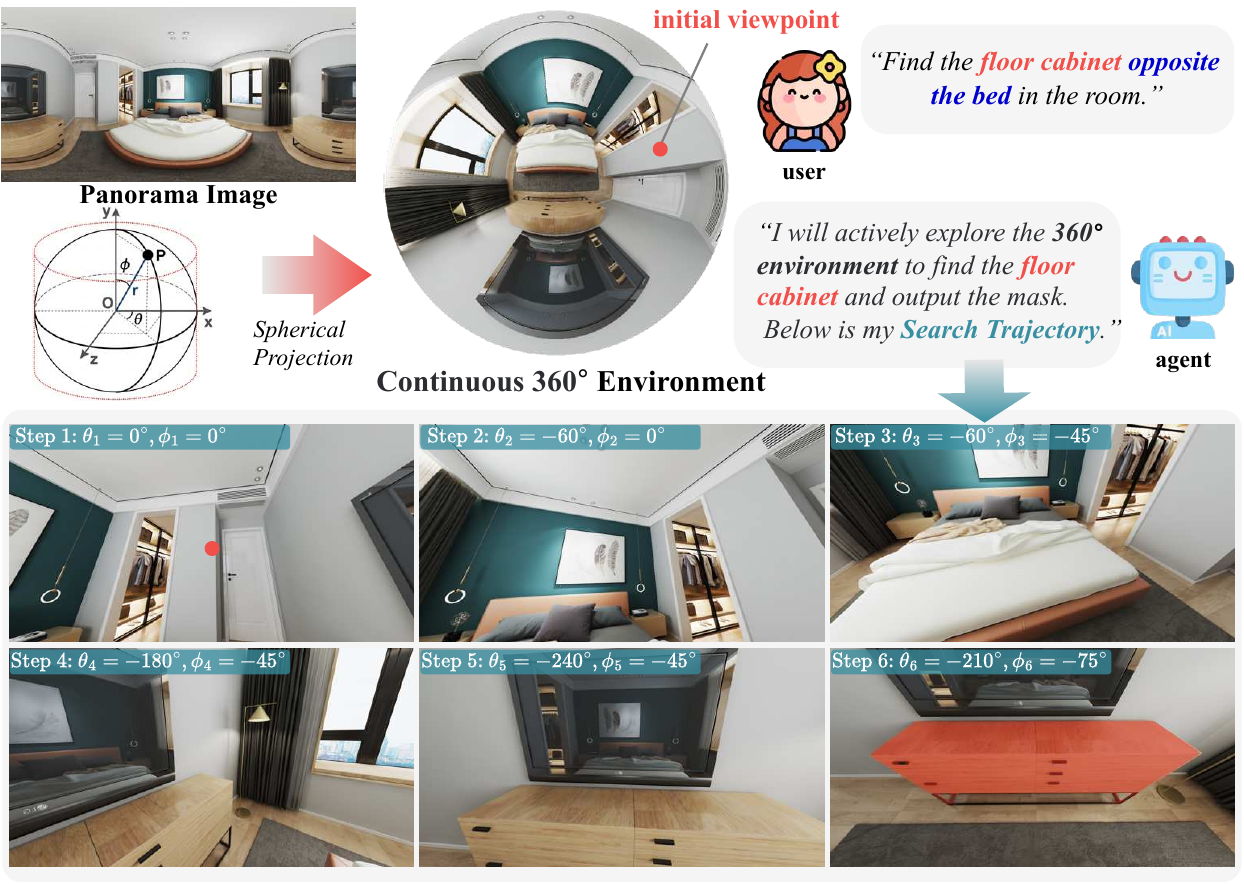}
    \caption{\textbf{Active Panoramic Referring Segmentation (APRS)} requires an agent to search for and segment a target (e.g., finding the ``floor cabinet '') within a continuous $360^{\circ}$ environment based on a user instruction. The bottom row illustrates a search trajectory: the agent adjusts its camera pose ($\theta, \phi$) to reason about spatial relations (e.g., ``opposite the bed'') and search the target for segmentation.}
   \label{fig:task}
\end{figure}

Transitioning from passive fixed-view observation to active multi-view perception presents fundamental challenges.
In $360^\circ$ environments, identifying a target necessitates reasoning over complex cross-view spatial relationships (e.g., ``the floor cabinet opposite the bed'') and performing active exploration.
To address these challenges, we introduce a novel task: \textbf{Active Panoramic Referring Segmentation~(APRS)}. In this setting, an agent is required to adjust its viewing direction~($\Delta\theta, \Delta\phi$) to explore the 360$^\circ$ environment, seeking the object specified by a user instruction for segmentation, as shown in~\cref{fig:task}.

However, applying existing perception methods to this task faces a dilemma.
On the one hand, explicit 3D modeling (e.g., point clouds~\cite{segpoint,ipdn,refmask3d,qin2025scenedesigner,shuai2025free,shuai2025free2,li2025anyi2v}, or Gaussian Splatting~\cite{refersplat,reasongrounder,3dgs}) resolves spatial queries by first reconstructing the entire 3D scene, which provides detailed spatial layouts but entails substantial computational and hardware costs.
On the other hand, conventional 2D methods~\cite{panovos,dense360,vqa360} that directly process flattened panoramic images introduce severe geometric distortions and wraparound boundaries, while those partitioning the $360^\circ$ scene into an isolated sequence of perspective views sacrifice global spatial context.

To address this dilemma, we construct the \textbf{APRS benchmark}, which leverages abundant and easily accessible panoramic images to simulate realistic $360^\circ$ egocentric exploration scenes.
It eliminates the need for resource-intensive 3D reconstructions while covering diverse indoor and outdoor environments.
Moreover, this benchmark features four distinct types of spatial referring expressions (EGO, UNIQ, ALLO, and MULTIHOP) to comprehensively evaluate agents' cross-view spatial reasoning ability across various difficulties.
We further provide comprehensive evaluation protocols for benchmarking.

Additionally, we propose \textbf{PanoSeeker}, a Vision-Language Agent designed for APRS.
To reduce redundant exploration and maintain global spatial context, we design  \textbf{EgoSphere}, an explicit spatial-visual memory, which integrates sequential local observations into a unified 360$^\circ$ equirectangular panorama (ERP) map.
Augmented by EgoSphere, PanoSeeker can perform efficient searching.
Once the target is found, the agent performs active viewpoint alignment and generates a segmentation mask.
Furthermore, we curate an expert-annotated search trajectory dataset with memory timelines for Supervised Fine-Tuning~(SFT). This is followed by Reinforcement Learning~(RL) using GRPO~\cite{deepseekmath}, which incorporates efficiency and terminal rewards to explicitly optimize PanoSeeker's exploration efficiency. Our main contributions are summarized as follows:

\begin{itemize}
    \item We introduce Active Panoramic Referring Segmentation (\textbf{APRS}), a novel task that requires an agent to actively explore $360^{\circ}$ environments to seek
    and segment targets based on language instructions. To support this, we construct the APRS benchmark, which incorporates four distinct types of spatial referring expressions and comprehensive evaluation protocols.

    \item We propose \textbf{PanoSeeker}, a memory-augmented agent that leverages an explicit spatial visual memory \textbf{EgoSphere} to facilitate efficient exploration.

    \item We contribute an expert-annotated search trajectory dataset with memory timelines for SFT. Additionally, we employ RL with efficiency and terminal rewards to explicitly optimize PanoSeeker's exploration efficiency.

    \item Extensive experiments demonstrate that PanoSeeker outperforms existing state-of-the-art baselines adapted for this task.
\end{itemize}

\section{Related Work}
\noindent \textbf{Referring Image Segmentation.}
Referring Image Segmentation~(RIS)~\cite{vlt, cris, res_survey} targets the segmentation of specific objects in an image according to a given textual description.
Early approaches~\cite{lstm-cnn,rmi,rrn,liu2024primitivenet} rely on CNNs to obtain visual features and LSTMs to encode the linguistic expressions, and then combine the two modalities via concatenation or other straightforward fusion operations to form a joint representation.
VLT~\cite{vlt,vltpami} is the first to bring the transformer into this task, recasting RIS as an attention problem in which language features serve as queries over the visual features and the segmentation result is obtained by decoding the transformer output.
Liu~\etal~\cite{gres} further extend the task to Generalized Referring Expression Segmentation (GRES), which accounts for both multi-target and empty-target cases.
More recently, Large Language Models (LLMs) and Vision Language Models (VLMs)~\cite{llama3,llava,deepseek-vl,deepseek-v3,internvl} have substantially advanced vision-language tasks through their strong common-sense reasoning ability, offering new opportunities for RIS.
Along this line, LISA~\cite{lisa} introduces the [SEG] token, which allows the model to handle expressions that demand complex reasoning and common-sense knowledge.
READ~\cite{read} treats similarity as reference points to direct where MLLMs should attend during interactive reasoning.
Seg-Zero~\cite{seg-zero} adopts a purely reinforcement-learning scheme (GRPO~\cite{deepseekmath}) to acquire the reasoning process from scratch, producing a reasoning chain prior to the final mask.
VisionReasoner~\cite{visionreasoner} presents a unified framework that addresses multiple visual perception tasks through reasoning within a single shared model.
Unlike these methods, which depend on static images captured from a fixed viewpoint, our PanoSeeker is able to actively perceive the continuous $360^{\circ}$ environment.

\noindent \textbf{Agent Memory.} 
Existing agent memory systems can be broadly categorized into flat-context and structured designs. 
MemGPT~\cite{memgpt} and MemoryOS~\cite{memoryos} adopt flat-context paradigms, which leverage paging or streaming mechanisms to extend memory capacity; however, their reliance on raw dialogue logs often introduces significant information redundancy and prohibitive retrieval overhead as histories expand.
To enhance semantic coherence and navigability, MemoryBank~\cite{memorybank} and A-Mem~\cite{a-mem} adopt the structured design to organize memory into hierarchical or graph-based representations. More recently, frameworks such as xMemory \cite{xmemory} have refined these structures by tailoring hierarchical retrieval to the dynamic nature of agentic workloads. 
Extending beyond textual modalities, vision-centric frameworks like VisMem \cite{vismem} incorporate short- and long-term latent visual memory into VLMs to enable robust long-term temporal reasoning.
Despite these advancements, existing history-based methods face a critical scalability bottleneck, as context length and query latency scale sharply with interaction history. 
To address this, we propose EgoSphere, a spatial visual memory that maps sequential observations onto a fixed-resolution $360^\circ$ panoramic canvas; this ensures a constant-length context regardless of the trajectory's length.

\noindent \textbf{Panoramic Vision and Embodied Referring.} 
Panoramic vision provides comprehensive $360^{\circ}$ contextual understanding, significantly advancing the ability of intelligent systems in navigation~\cite{refer360,reverie} and perception~\cite{dense360,vqa360,ai2025survey}. 
Early panoramic vision methods primarily focused on static images or videos with a $360^{\circ}$ Field-of-View (FoV), addressing challenges such as geometric distortion~\cite{2s-odis,osrt,tateno2018distortion}, representation learning~\cite{ai2025survey,su2017learning,ktn,shuai2024survey}, and multi-modal integration~\cite{dense360,vqa360,dongfang2025multimodal,shuai2026glyphprinter}. 
However, relying on static, fully observable panoramas diminishes the inherent need for active exploration and spatial reasoning. 
To address these limitations, recent embodied tasks (e.g., Refer360 ~\cite{refer360}, REVERIE~\cite{reverie}) integrate linguistic instructions into panoramic environments to inspire models' active perception and spatial reasoning ability. 
But they remain limited:
Refer360~\cite{refer360} relies on dense, step-by-step instructions that reduce the task to low-level instruction following, while REVERIE~\cite{reverie} provides multi-view observations at each step, turning search into a passive selection task. 
In contrast, our proposed Active Panoramic Referring Segmentation (APRS) requires the agent to adjust its viewing direction to explore the continuous $360^{\circ}$ environment, seeking the object specified by a user instruction for segmentation, thereby truly demonstrating the model's ability for active perception and spatial reasoning.

\section{The APRS Task and Benchmark}
\label{sec:task}
In this section, we formally define the Active Panoramic Referring Segmentation (APRS) task and describe the construction of our benchmark, including the dataset, environment setup, and evaluation protocols.

\subsection{Task Formulation}
\label{subsec:formulation}
Active Panoramic Referring Segmentation (APRS) requires an agent to search for and segment a target within a continuous $360^{\circ}$ environment $E$ based on a language instruction $I$. 
Starting from an initial orientation $(\theta_0, \phi_0)$ with a constrained Field of View (FoV), the agent iteratively updates its viewpoint $(\theta_t, \phi_t)$ to explore the environment. 
At each step $t$, the agent perceives a visual observation $V_t = \text{Proj}(E, \theta_t, \phi_t, \psi)$ via a perspective projection $\text{Proj}(\cdot)$ with an FoV $\psi$. 
Given $I$, the agent $\mathcal{F}$ maintains a memory state $M_t$ and predicts an action $a_t$:
\begin{equation}
    a_t, M_{t+1} = \mathcal{F}(V_t, I, M_t),
\end{equation}
where $M_{t+1}$ encapsulates the exploration history updated by the current observation $V_t$.
The action space includes relative movements $a_t = (\Delta \theta_t, \Delta \phi_t)$, denoting the direction of the viewpoint shift, and a termination signal that invokes the prediction of the segmentation mask $\mathcal{S}$ once the target is found.

\subsection{Environment and Dataset Construction}
\label{subsec:construction}
Unlike traditional Referring Image Segmentation (RIS)~\cite{res_survey,vltpami}, which assumes the target is always within the initial view, the APRS task necessitates active exploration to handle targets that are initially out of sight.
To this end, we construct a large-scale benchmark encompassing diverse indoor and outdoor scenes, complemented by complex referring expressions requiring spatial reasoning ability.

\noindent\textbf{Data Sources and Panoramic Simulation.} 
We collect high-resolution panoramas from 360-Indoor~\cite{360indoor}, PANDORA~\cite{pandora}, and SUN360~\cite{sun360}, resulting in 4,971 unique scenes.
This scale substantially surpasses existing language-guided 3D benchmarks such as ScanRefer~\cite{scanrefer} (comprising $\sim$800 indoor scenes) and offers greater diversity, spanning both indoor and outdoor environments.

To simulate a continuous $360^{\circ}$ environment, we employ gnomonic projection to extract local perspective views from equirectangular panoramas (ERPs). Given an orientation $(\theta, \phi)$ and a Field of View (FoV) $\psi=(\psi_h, \psi_v)$, we generate rectilinear views by back-projecting perspective coordinates onto a unit sphere and subsequently sampling from the ERP.
In our implementation, we set the horizontal FoV to $\psi_h = 120^{\circ}$ and the vertical FoV to $\psi_v = 90^{\circ}$, aligning with consumer-grade depth cameras.
More details are provided in the Appendix.

\noindent\textbf{Human-in-the-Loop Annotation Pipeline.}
We develop a multi-stage annotation pipeline that combines human expertise with foundation models~\cite{gpt5,sam3}. 
First, professional annotators annotate salient targets using bounding boxes. To encourage active exploration, initial camera orientations are directed away from the target, ensuring the target is initially out of the agent’s view.
Next, based on the spatial relationships between the initial camera orientation, the target, and other anchor objects, annotators are required to construct trajectory-centric spatial referring expressions. These expressions incorporate spatial relations, object attributes, and egocentric cues to guide the agent toward the target.
To achieve pixel-level segmentation, we leverage SAM-3~\cite{sam3} to generate initial masks, followed by manual refinement. 
Additionally, we employ VLMs (e.g., GPT-5.2~\cite{gpt5}) for semantic cross-checking to ensure the high reliability of our dataset.

\noindent\textbf{Taxonomy of Spatial Referring Expressions.} To evaluate diverse spatial reasoning ability, we categorize the spatial referring expressions into four types based on their linguistic and spatial complexity, as shown in Table~\ref{tab:referring_expression}:
\begin{itemize}[leftmargin=*, itemsep=3pt, topsep=3pt]
    \item \textbf{Egocentric (EGO)}: These instructions specify the target's location relative to the agent's current orientation. The primary challenge lies in mapping linguistic directional cues to precise spatial actions to effectively locate the target. (e.g., \textit{``Turn right to find the sofa.''})
    \item \textbf{Unique-Attribute (UNIQ)}: These instructions identify targets based on distinctive visual attributes (e.g., color or shape) to distinguish them from other objects of the same category within the $360^{\circ}$ scene. The core challenge involves fine-grained recognition and grounding of the specified instances across a panoramic view. (e.g., \textit{``The yellow floor lamp in the room.''})
    \item \textbf{Allocentric (ALLO)}: These expressions define the target's position relative to other anchor objects in the $360^{\circ}$ environment. The agent must interpret object-to-object spatial relationships to identify the correct target among potential distractors. (e.g., \textit{``The chair opposite the sofa.''})
    \item \textbf{Multi-hop (MULTIHOP)}: These instructions require complex reasoning by integrating both viewpoint adjustments and relative spatial cues. As the most challenging category, they demand sequential decomposition and multi-step reasoning. (e.g., \textit{``Turn around and find the chair next to the table.''})

\end{itemize}

\noindent\textbf{Supervised Fine-Tuning~(SFT) Dataset Construction.} 
Different from the initial APRS dataset annotation Phase, we recruit new annotators to generate exploration trajectories for fine-tuning the agent model.
Given an instruction $I$ and an initial camera orientation, annotators are required to find the target by adjusting the FoV based on camera pose parameters~($\theta, \phi$). 
We discretize the horizontal $\Delta \theta$ and vertical $\Delta \phi$ action space into multiples of $30^{\circ}$.
Horizontal actions include \textit{Small} ($30^{\circ}$), \textit{Medium} ($60^{\circ}$), \textit{Large} ($120^{\circ}$), and \textit{U-turn} ($180^{\circ}$); 
Vertical actions consist of \textit{Small} ($30^{\circ}$), \textit{Medium} ($60^{\circ}$), and \textit{Large} ($90^{\circ}$), facilitating a range from fine alignment to complete FoV shifts.
At each step, spatial coordinates~$(\theta_t, \phi_t)$, visual observations~$V_t$, and a memory~$M_t$ (detailed later in \cref{subsec:egosphere}) are recorded.
Trajectories exceeding the step limit or deviating significantly from the target are discarded and reassigned to different annotators.

\begin{table}[t]
\centering
\caption{\textbf{Taxonomy of Spatial Referring Expressions in the APRS Benchmark.} $O$ is the target object; $\{A\}$ denotes visual attributes; $R$ represents spatial relations; $v$ is the initial camera viewpoint; and $O_{anc}$ is the anchor object.}
\label{tab:referring_expression}
\setlength{\tabcolsep}{5.5pt}
\resizebox{\textwidth}{!}{ 
\setlength{\aboverulesep}{0pt}   
\setlength{\belowrulesep}{0pt}   
\begin{tabular}{lllc} 
\toprule
\rowcolor{myrowgray!70}
\textbf{Category} & \textbf{Formal Definition} & \textbf{Description} & \textbf{Number (\%)} \\ 
\midrule
\textbf{EGO}      & $find(O, \{A\}, R, v)$        & Relations relative to the initial camera viewpoint & 1,365~(18.4\%) \\ \addlinespace[0.5ex]
\rowcolor{myrowgray!50}
\textbf{UNIQ}     & $find(O, \{A\})$              & Objects with unique visual attributes (e.g., color) & 3,213~(43.3\%) \\ \addlinespace[0.5ex]
\textbf{ALLO}     & $find(O, \{A\}, R, O_{anc})$  & Relations relative to an external anchor object & 2,389~(32.2\%) \\ \addlinespace[0.5ex]
\rowcolor{myrowgray!50}
\textbf{MULTIHOP} & $EGO \circ ALLO$              & Multi-step reasoning via viewpoint and spatial cues & 453~(6.1\%) \\ 
\bottomrule
\end{tabular}
}
\end{table}
\noindent\textbf{Dataset Statistics.} The APRS dataset comprises 7,420 samples across 4,971 scenes, split into training and testing sets at a 7:3 ratio. On average, each scene contains 1.5 instructions spanning four difficulty levels: EGO, UNIQ, ALLO, and MULTIHOP. The average expression length is 7.38 words. These instructions incorporate diverse spatial prepositions (e.g., \textit{``opposite''}, \textit{``behind''}, and \textit{``adjacent to''}), necessitating spatial reasoning and active perception ability to search for and segment targets in the continuous 360$^\circ$ environments.

\subsection{Evaluation Metrics}
\label{subsec:metrics}

To comprehensively evaluate the performance of the APRS agent, we establish a multi-dimensional evaluation protocol comprising three dimensions: (i) task success, (ii) exploration efficiency, and (iii) segmentation quality.

\begin{itemize}[leftmargin=*, itemsep=3pt, topsep=3pt]

\item \textbf{Task Success.} 
We define the \textbf{Success Rate (SR)} to measure the agent's ability to search for and segment the referred target. An episode is considered successful if: (1) the search process terminates at step $T$; (2) the final viewpoint direction $(\theta_T, \phi_T)$ is within a geodesic distance threshold $\tau$ (calculated using the great-circle distance) from the target's ground-truth centroid $(\theta^*, \phi^*)$; and (3) the predicted mask $\mathcal{S}$ achieves an $\text{IoU} \ge 0.5$. Formally:
\begin{equation}
\begin{gathered}
    \text{SR} = \frac{1}{N} \sum_{i=1}^{N} \mathbb{I}(\text{dist}\{(\theta_{i,T}, \phi_{i,T}), (\theta_i^*, \phi_i^*)\} < \tau \land \text{IoU}_i \ge 0.5)
\end{gathered}
\label{eq:sr_dist}
\end{equation}
where $N$ is the total number of episodes and $\mathbb{I}(\cdot)$ is the indicator function. For each episode, the geodesic distance is defined as:
\begin{equation}
    \text{dist}\{(\theta_T, \phi_T), (\theta^*, \phi^*)\} = \arccos\left(\sin \phi_T \sin \phi^* + \cos \phi_T \cos \phi^* \cos(\theta_T - \theta^*)\right)
\end{equation}

\item \textbf{Exploration Efficiency.} 
Efficiency is crucial for active agents. We report the \textbf{Average Steps (AS)}, which represents the mean number of action steps taken before termination across successful episodes. Furthermore, we adapt \textbf{Success weighted by Path Length (SPL)} to the panoramic setting:
\begin{equation}
    \text{SPL} = \frac{1}{N} \sum_{i=1}^N S_i \cdot \frac{L_i}{\max(L_i, P_i)},
\end{equation}
where $S_i$ is the binary success indicator from the SR metric. $L_i$ represents the shortest geodesic distance from the initial orientation $(\theta_0, \phi_0)$ to the target's ground-truth centroid $(\theta^*, \phi^*)$, and $P_i$ is the cumulative great-circle distance traversed by the agent, defined as $P_i = \sum_{t=1}^{T} \text{dist}\{(\theta_t, \phi_t), (\theta_{t-1}, \phi_{t-1})\}$.

\item \textbf{Segmentation Quality.} 
Following standard Referring Image Segmentation benchmarks~\cite{res_survey, vlt, vltpami}, we evaluate the performance using \textbf{mean Intersection over Union (mIoU)}. For episodes where $\text{SR}=0$, the IoU is set to 0.
\end{itemize}

\section{Method}
\label{sec:method}
\begin{figure}[t]
    \centering
    \includegraphics[width=1\linewidth]{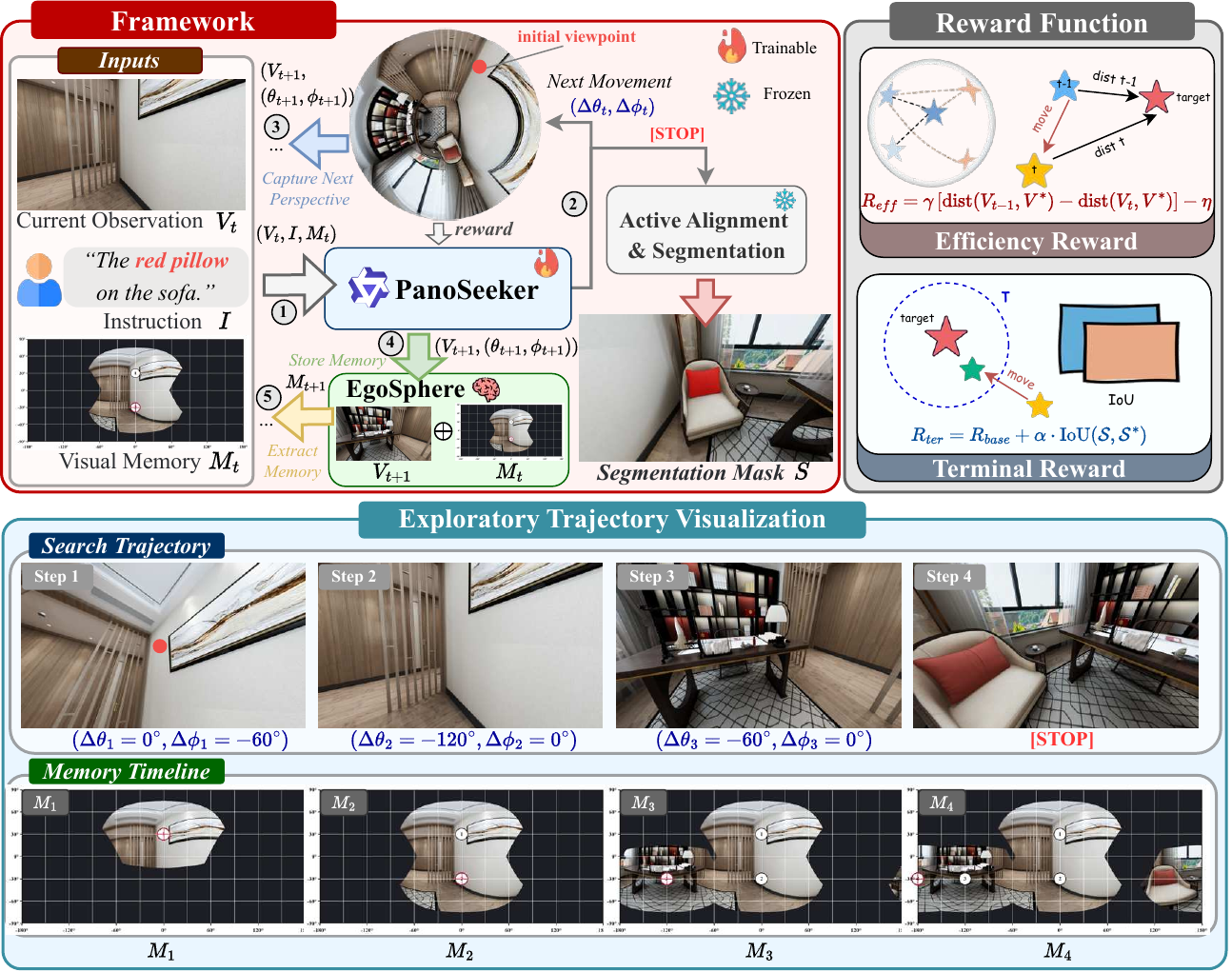}
    \caption{\textbf{Overview of the PanoSeeker framework.} PanoSeeker integrates a VLM with EgoSphere memory to predict movements $(\Delta\theta_t, \Delta\phi_t)$ from inputs $(V_t, I, M_t)$. Subsequently, an Active Alignment and Segmentation module generates the mask $S$. Exploration efficiency is guided by efficiency ($R_{\text{eff}}$) and terminal ($R_{\text{term}}$) rewards. Bottom: visualization of a search trajectory and its associated memory accumulation.}
    \label{fig:framework}
\end{figure}
We present \textbf{PanoSeeker}, a Vision-Language Agent integrated with a spatial visual memory \textbf{EgoSphere}, specifically designed for the APRS task. 
To ensure non-redundant active search and spatial reasoning, \textbf{PanoSeeker} is first trained on an expert-annotated trajectory dataset via Supervised Fine-Tuning (SFT), 
followed by Reinforcement Learning (RL) to explicitly optimize exploration efficiency. 
Once the target is found, an \textbf{Active Alignment and Segmentation} module is employed for final viewpoint refinement and pixel-level segmentation.

\subsection{EgoSphere: Spatial Visual Memory}
\label{subsec:egosphere}
A primary challenge in APRS is maintaining global spatial consistency under a constrained FoV. Existing memory methods often rely on textual logs or isolated perspective crops—which fail to capture the geometric continuity of a 360$^\circ$ environment and lack trajectory history, leading to redundant exploration (i.e., dead loops).
To bridge this gap, we introduce \textbf{EgoSphere},  an explicit memory which progressively maps sequential local observations into a geometrically consistent, annotated 360$^\circ$ exploration map to ensure targeted, non-redundant search.

\noindent\textbf{Progressive Canvas Construction.} 
We instantiate the spatial memory as an Equirectangular Panorama (ERP) canvas $M \in \mathbb{R}^{H \times W \times 3}$, initialized as a zero-valued tensor. 
At each step $t$, the agent receives a perspective FoV $V_t$ at orientation $(\theta_t, \phi_t)$. 
To integrate this local observation, we back-project $V_t$ onto a unit sphere and map it to the ERP coordinates:
\begin{equation}
    M_t = M_{t-1} \oplus \text{Proj}^{-1}(V_t, \theta_t, \phi_t, \psi),
\end{equation}
where $\text{Proj}^{-1}(\cdot)$ denotes the inverse gnomonic projection and $\oplus$ signifies the pixel-wise update. 
This formulation ensures $M_t$ serves as a cumulative visual record, where unexplored regions are explicitly represented as zero-valued frontiers (see Appendix for more gnomonic projection details).

\noindent\textbf{Spatial-Aware Visual Prompting.} 
To enable spatial reasoning, we augment $M_t$ with explicit geometric cues that transform the canvas into a structured navigation map. Specifically, we incorporate three types of \textbf{visual prompts}: 
(1) a \textbf{central crosshair} that highlights the current FoV center to facilitate egocentric-to-allocentric alignment; 
(2) a \textbf{latitude-longitude grid} at $30^\circ$ intervals, which serves as a global scale for precise angular estimation; and
(3) the \textbf{exploration trajectory} $\{(\theta_1, \phi_1), \dots, (\theta_{t-1}, \phi_{t-1})\}$, rendered as directed paths to track visited regions and prevent redundant scanning.
By integrating the recorded memory $M_t$ with the current view $V_t$, the agent effectively converts the search problem into a visual ``fill-in-the-blank'' task on the $360^\circ$ canvas.

\subsection{PanoSeeker}
The PanoSeeker serves as the spatial reasoning core, leveraging a Vision-Language Model (VLM)~\cite{qwen3vl} to guide camera orientation control. 
This module processes a multi-modal input consisting of three key streams: (1) the current local observation $V_t$; (2) the annotated spatial memory map $M_t$ from EgoSphere; and (3) the instruction $I$. 
By jointly attending to the local view (for object detail) and the global map (for path planning), PanoSeeker addresses the search problem by predicting the next optimal action $a_t \in \{(\Delta \theta_t, \Delta \phi_t), \texttt{[STOP]}\}$.

\noindent\textbf{Supervised Fine-Tuning (SFT) Stage.} 
To empower the VLM~\cite{qwen3vl} with active perception ability, we fine-tune PanoSeeker using an expert-annotated trajectory dataset. 
This stage explicitly trains PanoSeeker to predict the optimal viewpoint adjustment based on the current context. The objective function is defined as:
\begin{equation}\label{eq:sft}
\mathcal{L} = - \mathbb{E}_{\mathcal{T} \sim \mathcal{D}_{\text{expert}}} \sum_{t=1}^{T} \log p_{\theta}(a_t \mid V_t, M_t, I),
\end{equation}
where each expert trajectory  $\mathcal{T} = \{(V_t, M_t, a_t, I)\}_{t=1}^T$ is sampled from the expert dataset $\mathcal{D}_{\text{expert}}$ (detailed in \cref{subsec:construction}). The loss minimizes the negative log-likelihood of the expert action $a_t$ given the learned parameters $\theta$ of PanoSeeker, thereby learning the policy for efficient, non-redundant search.

\noindent\textbf{Reinforcement Learning (RL) Stage.} 
To enhance search efficiency, we fine-tune PanoSeeker using Group Relative Policy Optimization (GRPO)~\cite{deepseekmath}, which optimizes the policy through relative rewards within sampled groups without a separate critic. 
To formulate this reinforcement learning process, we first define the reward mechanism to guide the agent's behavior. 
For a navigation trajectory comprising states $s_t = (V_t, M_t, I)$ of length $T$, the overall cumulative reward is defined as $r = \sum_{t=1}^{T-1} R_{eff} + R_{term}$. 
Specifically, the \textbf{efficiency reward} $R_{eff} = \gamma [\text{dist}(V_{t-1}, V^*) - \text{dist}(V_t, V^*)] - \eta$ is computed at each step to encourage the agent to follow the shortest geodesic path while penalizing redundant exploration. 
Upon generating the \texttt{[STOP]} token at step $T$, the \textbf{terminal reward} $R_{term}$ assigns a success bonus $R_{base} + \alpha \cdot \text{IoU}$ if the agent successfully finds the target ($\text{dist}(V_T, V^*) < \tau$), and imposes a penalty $-R_{pen}$ otherwise.
Given the cumulative reward $r$ evaluated for each trajectory in a sampled group, we maximize the following GRPO objective:

{\small
\begin{equation}
\label{eq:grpo}
\begin{aligned}
\mathbb{E}_{\substack{s_t \sim \mathcal{D}, \\ \{a_{t,i}\}_{i=1}^{G} \sim p_{\theta_{\text{old}}}}}
\Bigg[ \frac{1}{G} \sum_{i=1}^{G} \Bigg( &\min \Bigg( \frac{p_\theta(a_{t,i} \mid s_t)}{p_{\theta_{\text{old}}}(a_{t,i} \mid s_t)}A_i, \text{clip}\Bigg(\frac{p_\theta(a_{t,i} \mid s_t)}{p_{\theta_{\text{old}}}(a_{t,i} \mid s_t)}, 1-\epsilon, 1+\epsilon\Bigg)A_i \Bigg) \\
&- \beta\, \mathbb{D}_{\text{KL}}\!\left(p_\theta \,\|\, p_{\text{ref}}\right) \Bigg) \Bigg],
\end{aligned}
\end{equation}
}

\noindent where $p_{\text{ref}}$ is the SFT reference model, $G$ is the group size, and $\epsilon$ is the clipping parameter. The advantage $A_i$ for each action $a_{t,i}$ is computed by normalizing its reward $r_i$ within the group: $A_i = {(r_i - \text{mean}(\{r_g\}))}/{\text{std}(\{r_g\})}$. $\beta$ weights the KL-divergence $\mathbb{D}_{\text{KL}}$ for regularization.

\subsection{Active Alignment and Segmentation}
\label{subsec:alignment}
Upon receiving the \texttt{[STOP]} signal, PanoSeeker predicts a bounding box $\mathcal{B}$ by leveraging its inherent VLM~\cite{qwen3vl} grounding ability.
To address the challenges of boundary truncation and occlusion—which can lead to unstable mIoU metric evaluation—we implement an active alignment phase. Taking $\mathcal{B}$ as a prompt, we augment SAM-3~\cite{sam3} to perform concurrent center-seeking, tracking, and segmentation. By applying a final angular adjustment $(\Delta \theta^*, \Delta \phi^*)$, the target's centroid is realigned to the optical center. This ensures the generation of a stable and accurate segmentation mask $\mathcal{S}$ for reliable performance evaluation.

\section{Experiments}
\label{sec:experiments}
\subsection{Implementation Details}
\noindent \textbf{Datasets and Environment.} 
The APRS dataset comprises 7,420 samples across 4,971 scenes, partitioned 7:3 for training and testing.
To provide a comprehensive evaluation, we adopt four primary metrics: (1) Success Rate (SR) to measure target localization accuracy; (2) Average Steps (AS) to assess search efficiency; (3) Success weighted by Path Length (SPL) to evaluate trajectory optimality; and (4) mean Intersection over Union (mIoU) to quantify segmentation quality. 
Each episode is limited to a maximum of $T=20$ search steps. 
Each visual observation is rendered with a horizontal Field-of-View ($\psi_h$) of $120^\circ$ and a vertical Field-of-View ($\psi_v$) of $90^\circ$, at a resolution of $1024 \times 768$ pixels.

\noindent \textbf{Network Architecture: } 
We employ Qwen3-VL-8B-Instruct~\cite{qwen3vl} as the backbone for our PanoSeeker, benefiting from its superior reasoning and grounding ability for the APRS task. 
Specifically, the model is fine-tuned using a LoRA~\cite{lora} adapter with a rank $r=64$, a scaling factor $\alpha=64$, and a dropout rate of $0.05$. 
Furthermore, the EgoSphere memory is maintained at a resolution of $1024 \times 512$, while SAM-3~\cite{sam3} is used for final segmentation tasks.

\noindent \textbf{Training Details: } 
We implement our model using the DeepSpeed ZeRO-2 framework~\cite{deepspeed} and conduct training on $4 \times$ NVIDIA RTX A6000 GPUs (48GB) with FP16 precision. 
For Supervised Fine-Tuning (SFT), we employ the AdamW~\cite{adamw} optimizer with a learning rate of $2 \times 10^{-4}$ and a weight decay of $0.1$. A cosine annealing scheduler is applied with a $3\%$ warmup over $3$ epochs. The training is performed with a per-GPU batch size of $1$ and a gradient accumulation step of $4$, resulting in an effective batch size of $16$, with a sequence length limit of $4,096$ tokens. 
In the Reinforcement Learning stage, we leverage the GRPO~\cite{deepseekmath} ($G=8, \epsilon=0.2$) for $6,000$ steps to explicitly optimize exploration efficiency.

\begin{table}[t]
\centering
\caption{\textbf{Quantitative comparison on the APRS benchmark.} We compare PanoSeeker with static RIS (processing the entire panorama), heuristic scanning (Upper $\rightarrow$ Equator $\rightarrow$ Lower), and active VLM agents. PanoSeeker achieves a significant leap in exploration efficiency (SPL and AS) thanks to its EgoSphere spatial memory and GRPO-optimized~\cite{deepseekmath} policy. The best results are in \textbf{bold}.}
\label{tab:main_results_final}
\setlength{\tabcolsep}{6pt}
\resizebox{\linewidth}{!}{
\begin{tabular}{lcccccc}
\toprule
\textbf{Method} & \textbf{Venue} & \textbf{Memory} & \textbf{SR (\%) $\uparrow$} & \textbf{AS $\downarrow$} &  \textbf{SPL $\uparrow$} & \textbf{mIoU (\%) $\uparrow$} \\
\midrule
\multicolumn{7}{l}{\textbf{\textit{a. Static Methods (Direct Panorama Input)}}} \\
\rowcolor{gray!10}
VLT~\cite{vlt} & \pub{ICCV'21}  & \xmark & 46.7 & 1.0$^\ast$ & - &  34.7 \\
CRIS~\cite{cris} & \pub{CVPR'22}  & \xmark & 55.4 & 1.0$^\ast$ & - &  39.2 \\
\rowcolor{gray!10}
LISA~\cite{lisa} & \pub{CVPR'24} & \xmark & 64.1 & 1.0$^\ast$ & - &  44.5 \\
SAM4MLLM~\cite{sam4mllm} & \pub{ECCV'24}  & \xmark & 62.3 & 1.0$^\ast$ & - &  46.2 \\
\rowcolor{gray!10}
VisionReasoner~\cite{visionreasoner} & \pub{ICLR'26}  & \xmark & 66.2 & 1.0$^\ast$ & - &  47.7 \\
\midrule
\multicolumn{7}{l}{\textbf{\textit{b. Heuristic Methods (Pre-defined Scanning, Max 9 steps)$^\dagger$}}} \\
VLT~\cite{vlt} & \pub{ICCV'21} & \xmark & 28.4 & 4.6 & 0.14 & 22.1 \\
\rowcolor{gray!10}
CRIS~\cite{cris} & \pub{CVPR'22} & \xmark & 32.1 & 4.4 & 0.17 & 26.4 \\
LISA~\cite{lisa} & \pub{CVPR'24} & \xmark & 42.3 & 5.2 & 0.22 & 34.1 \\
\rowcolor{gray!10}
SAM4MLLM~\cite{sam4mllm} & \pub{ECCV'24} & \xmark & 40.5 & 5.3 & 0.20 & 32.5 \\
VisionReasoner~\cite{visionreasoner} & \pub{ICLR'26} & \xmark & 49.5 & 4.9 & 0.26 & 39.9 \\
\midrule
\multicolumn{7}{l}{\textbf{\textit{c. VLM-based Agents (Active Exploration)}}} \\
\rowcolor{gray!10}
Qwen3-VL-30B-A3B-Thinking ~\cite{qwen3vl} & \pub{ArXiv'25} & Text Log & 62.9 & 8.0 & 0.29 & 51.0 \\
Gemini-3-Flash~\cite{gemini} & - & Text Log & 64.9 & 6.8 & 0.31 & 51.4 \\
\rowcolor{gray!10}
GPT-5.2~\cite{gpt5} & - & Text Log & 69.1 & 6.2 & 0.36 & 53.2 \\
\rowcolor{cyan!10}
\textbf{PanoSeeker (Ours)} & - & \textbf{EgoSphere} & \textbf{75.4} & \textbf{4.8} & \textbf{0.57} & \textbf{55.8} \\
\bottomrule
\multicolumn{7}{l}{$^\ast$ \textit{Static methods process the panorama in a single pass (AS$=$1); SPL is therefore not applicable ($-$).}} \\
\multicolumn{7}{l}{$^\dagger$ \textit{Heuristic scanning follows a fixed order: Upper Hemis. $\rightarrow$ Equator $\rightarrow$ Lower Hemis.}} \\
\end{tabular}
}
\end{table}
\subsection{APRS Benchmark Results}
\label{subsec:main_results}

In Table~\ref{tab:main_results_final}, we report the quantitative results on our APRS benchmark, comparing PanoSeeker with previous state-of-the-art methods under three search paradigms: 
(1) \textit{Static Methods}, which directly process the $360^{\circ}$ equirectangular panorama (ERP) for referring image segmentation~(RIS) in a single forward pass; 
(2) \textit{Heuristic Methods}, which follow a predefined scanning paradigm traversing the upper, middle (equatorial), and lower spheres in order, ensuring complete $360^{\circ}$ coverage of the space, where the search terminates once the model predicts a segmentation mask; and
(3) \textit{VLM-based Agents}, which serve as active controllers using task-specific prompts and textual log memory.

\noindent\textbf{Quantitative Comparison.}
As shown in Table~\ref{tab:main_results_final}, PanoSeeker achieves the best results across all metrics. 
Static methods struggle with panoramic image distortion and fail to recognize spatial referring expressions that are hard to identify in a single flat panorama. 
While heuristic scanning ensures coverage, its fixed path lacks reasoning, leading to many extra steps and a low 0.26 SPL. 
In contrast, active agents are more suitable for this task because they can reason during exploration. 
PanoSeeker achieves 75.4\% SR and 55.8\% mIoU, improving SPL by $\sim$58\% over GPT-5.2 (0.57 vs. 0.36) with only 4.8 steps. These gains are attributed to our EgoSphere's explicit spatial visual memory for path planning and the GRPO-optimized~\cite{deepseekmath} policy that enhances exploration efficiency.

\begin{figure}[t]
    \centering
    \includegraphics[width=1\linewidth]{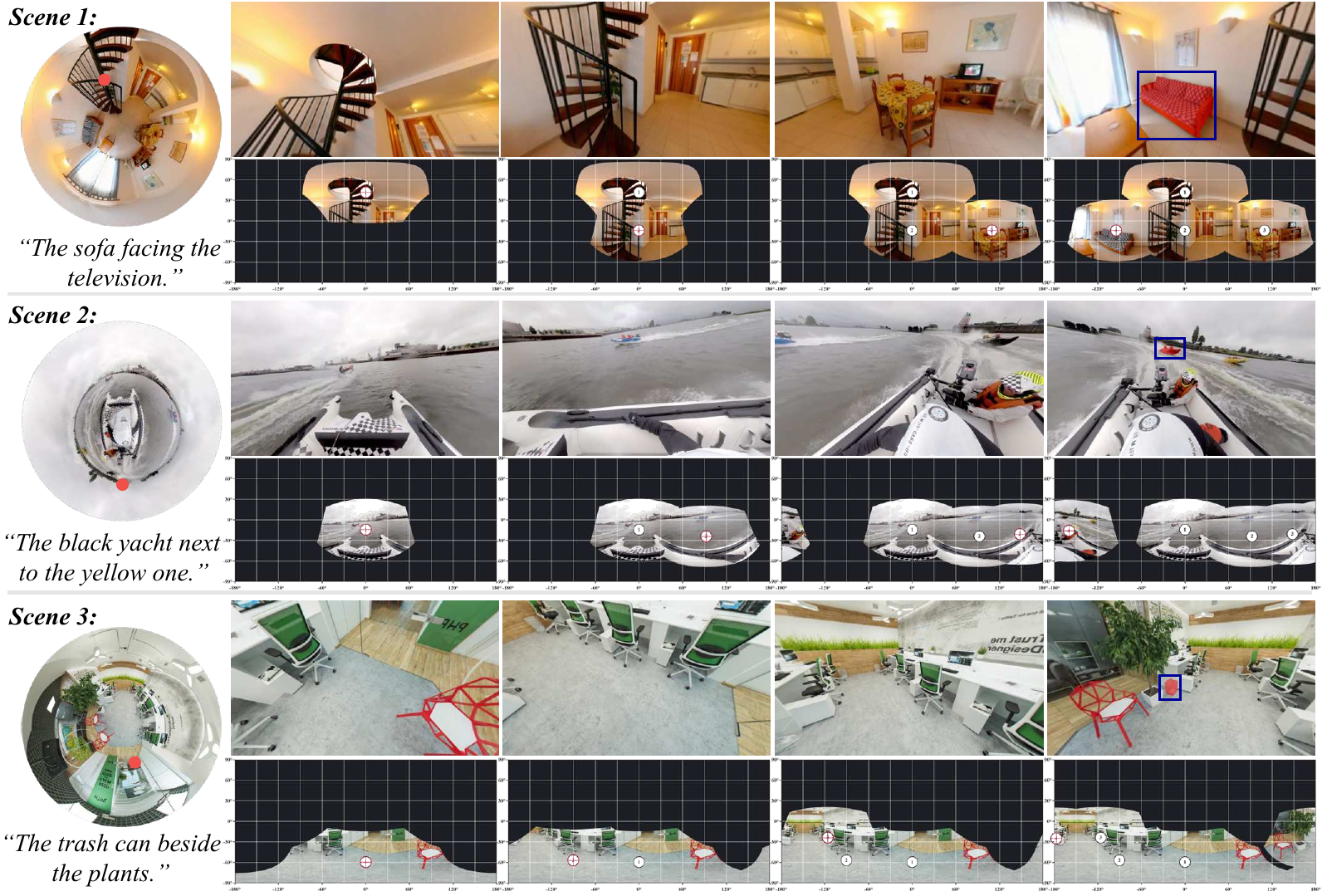}
    \caption{\textbf{Visualization Results.} \textbf{\textcolor{myRed}{Red dots}} denote initial viewpoints; \textbf{\textcolor{myBlue}{blue boxes}} highlight final targets found by PanoSeeker. While these visualizations only show 4-step trajectories, extended scenarios can be found in the Appendix.}
   \label{fig:qualitative}
\end{figure}

\noindent\textbf{Visualization.} 
\cref{fig:qualitative} shows PanoSeeker's search trajectories and how the EgoSphere is built step-by-step. 
Starting from the initial viewpoints (red dots), the agent iteratively adjusts its viewing direction~($\Delta\theta, \Delta\phi$) to explore the $360^{\circ}$ environment for seeking and segmenting targets specified by language instructions.
The EgoSphere progressively maps sequential local observations into a $360^{\circ}$ map, helping the agent perceive the whole scene and improve exploration efficiency.
These cases are only selected from episodes with the trajectory length of 4; more complex scenarios with longer trajectories are provided in the Appendix.

\subsection{Ablation Study}
\label{subsec:ablation}
We conduct ablation experiments to verify the effectiveness of all components. All variants use Qwen3-VL-8B~\cite{qwen3vl} as the backbone. All results are processed by the Active Alignment and Segmentation module to ensure stable evaluation.
\begin{table}[h]
\centering
\caption{Ablation of \textbf{PanoSeeker components}. (a) is a zero-shot baseline using task-specific prompts. We incrementally build upon it to reach our full framework (d).}
\label{tab:ablation_main}
\setlength{\tabcolsep}{8pt}
\resizebox{1\linewidth}{!}{
\begin{tabular}{lcccccccc}
\toprule
\textbf{Variant} & \textbf{Training} & \textbf{EgoSphere} & \textbf{GRPO~\cite{deepseekmath}} & \textbf{SR (\%) $\uparrow$} & \textbf{AS $\downarrow$} & \textbf{SPL $\uparrow$} & \textbf{mIoU (\%) $\uparrow$} \\
\midrule
\rowcolor{myrowgray!50}
(a) Zero-shot                  & None   &    \xmark        &  \xmark          & 41.6 & 12.4 & 0.14 & 28.5 \\
(b) + Memory                   & None   & \cmark &       \xmark     & 56.5 & 8.7  & 0.26 & 39.9 \\
\rowcolor{myrowgray!50}
(c) + SFT                      & SFT    & \cmark &     \xmark       & 70.3 &  6.2 & 0.40 & 50.7 \\
\rowcolor{cyan!10}
(d) + RL Strategy \textbf{(Full)} & SFT+RL & \cmark & \cmark & \textbf{75.4} & \textbf{4.8} & \textbf{0.57} & \textbf{55.8} \\
\bottomrule
\end{tabular}
}
\end{table}
\begin{table}[h]
\centering
\caption{Comparison of different \textbf{memory architectures} in a zero-shot setting. All models are prompted to perform active searching. ``Visual Hist.'' denotes the format of stored historical visual observations. ``ERP'' denotes Equirectangular Panorama format.}
\label{tab:ablation_memory}
\setlength{\tabcolsep}{10.0pt}
\resizebox{1\textwidth}{!}{
\begin{tabular}{lcccccc}
\toprule
\textbf{Memory Variant} & \textbf{Text Log} & \textbf{Visual Hist.} & \textbf{SR (\%) $\uparrow$} & \textbf{AS $\downarrow$} & \textbf{SPL $\uparrow$} & \textbf{mIoU (\%) $\uparrow$} \\
\midrule
\rowcolor{myrowgray!50}
(a) Baseline (Zero-shot)& \xmark & \xmark & 41.6 & 12.4 & 0.14 & 28.5 \\
(b) + Textual Log       & \cmark & \xmark & 50.3 & 9.9  & 0.21 & 35.7 \\
(c) + Visual Buffer     & \xmark & 5 frames & 46.8 & 10.5 & 0.17 & 32.0 \\
\rowcolor{myrowgray!50}
(d) + Hybrid Buffer     & \cmark & 5 frames & 43.6 & 11.0 & 0.15 & 30.6 \\
\rowcolor{cyan!10}
(e) + \textbf{EgoSphere (Ours)} & \xmark & \textbf{1 ERP canvas} & \textbf{56.5} & \textbf{8.7} & \textbf{0.26} & \textbf{39.9} \\
\bottomrule
\end{tabular}
}
\end{table}

\noindent \textbf{Component Effectiveness.} 
Table~\ref{tab:ablation_main} shows the contribution of each component to the full PanoSeeker framework. The zero-shot baseline (a) fails to navigate the $360^\circ$ environment effectively, resulting in a high average step (AS=12.4) and low SPL (0.14) due to frequent ``dead loops.'' Integrating EgoSphere (b) significantly improves efficiency, boosting the SPL to 0.26 by providing a persistent visual record that prevents redundant exploration. While the SFT stage (c) further improves the SR to 70.3\% by learning experts' search trajectories, the addition of GRPO-based RL (d) achieves the best performance. Specifically, the RL strategy optimizes search trajectories for efficiency, reducing the average steps to 4.8 and reaching the highest SPL of 0.57. This demonstrates that while memory provides the necessary spatial context and historical trajectory, reinforcement learning is key to achieving optimal, non-redundant search paths.

\noindent \textbf{Analysis of Memory Architectures.} 
Table~\ref{tab:ablation_memory} compares various memory strategies in a zero-shot setting. We observe that textual logs (b), visual buffers (c), and hybrid approaches (d) lack sufficient spatial reasoning abilities. Even when past observations are recorded, these sequence-based methods struggle to maintain global context; notably, the hybrid buffer (d) even shows decreased performance due to multimodal information conflicts in the input (43.6 vs. 46.8 SR). In contrast, EgoSphere (e) uses geometric projection to unify local views into a continuous $360^{\circ}$ representation, significantly outperforming the next-best variant (Textual Log) by 6.2\% in SR and 0.05 in SPL. This highlights that for the APRS task, a geometrically consistent map is far more effective for spatial awareness than simply accumulating temporal logs or isolated observation frames.

\section{Conclusion}
\label{sec:conclusion}
In this paper, we introduce Active Panoramic Referring Segmentation (APRS), a novel task that requires an agent to actively explore $360^\circ$ environments for target search and segmentation based on instructions. 
To address this, we propose PanoSeeker, a memory-augmented agent equipped with EgoSphere—an explicit spatial-visual memory that integrates sequential observations into a unified panoramic representation. 
By combining Supervised Fine-Tuning with Reinforcement Learning, we explicitly optimize the agent's active perception and spatial reasoning ability. 
Extensive experiments show that PanoSeeker outperforms existing adapted state-of-the-art baselines, establishing APRS as a scalable, cost-effective benchmark for advancing active perception in Embodied AI.

\bibliographystyle{unsrtnat}
\bibliography{main}

\end{document}